\DocumentMetadata{
    pdfversion=2.0,pdfstandard=ua-2,
    testphase={phase-III,firstaid,math,title}
}

\documentclass[sigconf,screen]{acmart-tagged} 
\usepackage[utf8]{inputenc}
\usepackage{newunicodechar}
\newunicodechar{（}{(}
\newunicodechar{，}{,}

\AtBeginDocument{%
  }

\copyrightyear{2026}
\acmYear{2026}
\setcopyright{cc}
\setcctype{by}
\acmConference[HRI '26]{Proceedings of the 21st ACM/IEEE International Conference on Human-Robot Interaction}{March 16--19, 2026}{Edinburgh, Scotland Uk}
\acmBooktitle{Proceedings of the 21st ACM/IEEE International Conference on Human-Robot Interaction (HRI '26), March 16--19, 2026, Edinburgh, Scotland Uk}
\acmPrice{}
\acmDOI{10.1145/3757279.3785565}
\acmISBN{979-8-4007-2128-1/2026/03}

\usepackage{stfloats}
\usepackage{longtable}
\usepackage{booktabs}
\usepackage{array}
\usepackage[table]{xcolor} 
\usepackage{multirow}
\usepackage{float}
\usepackage{arydshln}
\usepackage{booktabs} 
\usepackage{tabularx}
\begin{document}
\title[Designing Persuasive Social Robots for Health Behavior Change: A Systematic Review]{Designing Persuasive Social Robots for Health Behavior Change: A Systematic Review of Behavior Change Strategies and Evaluation Methods}

\author{Jiaxin Xu}
\affiliation{
  \institution{Eindhoven University of Technology}
  \city{Eindhoven}
  \country{Netherlands}}
\email{j.xu2@tue.nl}

\author{Chao Zhang}
\affiliation{
  \institution{Eindhoven University of Technology}
  \city{Eindhoven}
  \country{Netherlands}}
\email{c. zhang.5@tue.nl}

\author{Raymond H. Cuijpers}
\affiliation{
  \institution{Eindhoven University of Technology}
  \city{Eindhoven}
  \country{Netherlands}}
\email{r.h.cuijpers@tue.nl}

\author{Wijnand A. IJsselsteijn}
\affiliation{
  \institution{Eindhoven University of Technology}
  \city{Eindhoven}
  \country{Netherlands}}
\email{w.a.ijsselsteijn@tue.nl}

\begin{abstract}
Social robots are increasingly applied as health behavior change interventions, yet actionable knowledge to guide their design and evaluation remains limited. This systematic review synthesizes (1) the behavior change strategies used in existing HRI studies employing social robots to promote health behavior change, and (2) the evaluation methods applied to assess behavior change outcomes. Relevant literature was identified through systematic database searches and hand searches. Analysis of 39 studies revealed four overarching categories of behavior change strategies: coaching strategies, counseling strategies, social influence strategies, and persuasion-enhancing strategies. These strategies highlight the unique affordances of social robots as behavior change interventions and offer valuable design heuristics. The review also identified key characteristics of current evaluation practices, including study designs, settings, durations, and outcome measures, on the basis of which we propose several directions for future HRI research.
\end{abstract}



\maketitle

\section{Introduction}
HRI user studies are gradually moving beyond conventional concerns with usability or user acceptance to explore how robots can influence human behavior \cite{weiss2009usus}. Within this evolving landscape, a distinct research area known as \textit{persuasive social robots} \cite{ham2011} has emerged, focusing on the design of social robots that can influence users’ attitudes or behaviors in specific contexts \cite{liu2022}. A promising application of persuasive social robots is the promotion of health behavior change, such as motivating people to exercise more \cite{89_winkle2019,127_rea2021}, eat healthier \cite{604_baroni2014}, or adhere to medication regimens \cite{648_broadbent2014}. Although still in its early stages, this application has the potential to complement existing digital health interventions and help address pressing public health challenges linked to unhealthy lifestyles \cite{H2_giorgi2025,stanojevic12digital}.

Compared to conventional persuasive technologies (e.g., mobile apps and websites), social robots offer unique advantages for supporting health behavior change. Most notably, they are physically present with users \cite{H6_casas2021}. Such presence provides a strong social cue that can lead people to take advice more seriously \cite{ham2011}. Empirical studies show that people are more likely to comply with persuasion from physically embodied robots than from disembodied agents \cite{fischer2012levels,li2015benefit,wainer2007}. Beyond physical presence, social robots combine rich social capabilities (e.g., verbal and nonverbal communication) with animate qualities (e.g., human- or animal-like appearance and movement) \cite{laban2024}, both of which can elicit stronger compliance responses \cite{ham2011}. Equipped with sensors and processors, they can also identify activity patterns and health status and provide real-time feedback, enabling more personalized coaching \cite{stanojevic12digital, spitale2022}. These affordances make social robots a promising extension of persuasive technologies that warrants further exploration.

However, designing social robots as effective behavioral interventions is not straightforward. It requires an understanding of complex \textit{behavior change strategies}, that is, theoretically valid techniques capable of influencing behavior (e.g., goal setting, self-monitoring) \cite{michie2011strengthening}. These strategies form the foundation of any behavioral intervention by linking broad intervention aims (e.g., increasing physical activity) with specific design elements (e.g., visualizations, prompts, or texts) \cite{mohr2014}. In conventional persuasive technologies, considerable progress has been made in mapping which strategies can be implemented, how they can be instantiated through different technical features, and which are potentially most effective, as demonstrated by numerous reviews \cite{duff2017,lu2025,aldenaini2023} and meta-analyses \cite{webb2010,almutairi2023}. In contrast, the use of social robots for health behavior change remains limited \cite{liu2022}. At present, we lack a foundational understanding of what behavior change strategies mean in HRI contexts and how they can be translated into social robot design.

Another well-recognized challenge in designing behavioral intervention technologies is evaluating behavior change outcomes \cite{klasnja2011}. Health behavior change is a long-term, non-linear process with frequent relapses \cite{marcus2000}. Demonstrating that a technology can genuinely support this process requires rigorous study designs, extended observation periods, and valid behavioral measures \cite{klasnja2011}. These challenges are even more pronounced in HRI, where social robots are still technically fragile, costly, and difficult to deploy in real-world settings \cite{261_ruf2020}. Under these constraints, HRI researchers must plan their evaluations with particular precision and caution. A systematic review of current evaluation methods is therefore needed to support informed methodological choices. 

Together, the need to understand both behavior change strategies and evaluation methods motivates this review. We systematically searched, coded, and analyzed existing empirical HRI studies in which social robots were used to promote health behavior change. Our first aim is to identify \textbf{(RQ1)} which behavior change strategies have been implemented in the design of these robots, and our second aim is to examine \textbf{(RQ2)} how behavior change outcomes have been evaluated. This dual focus enables us to comprehensively assess the state of the field and provide actionable insights for designing and evaluating health-promoting social robots.

\section{Related work}
\subsection{Persuasive social robots}
Persuasion, a concept rooted in psychology, originally refers to interpersonal communication that shapes others’ independent judgments and actions \cite{simons2001persuasion}. The HCI field adopted this idea decades ago, introducing the term \textit{persuasive technology} to describe technologies intentionally designed to influence users’ attitudes or behaviors \cite{fogg2002}. At the time, screen-based platforms such as mobile phones, tablets, and computers were seen as especially promising due to their broad accessibility, low cost, and ability to combine interpersonal and mass communication strategies \cite{oinas2008,klasnja2011}.

Recently, researchers have begun to examine the potential of social robots as persuasive technologies, and several reviews have mapped this emerging field. Liu et al. \cite{liu2022} reviewed 60 experimental studies on persuasive social robots and identified five design factors most frequently investigated: modality, interaction, social character, context, and persuasive strategies. Saunderson and Nejat \cite{saunderson2019robots} provided a narrative review, summarizing how robots’ nonverbal behaviors can influence users’ cognitive framing, emotions, behavioral responses, and task performance. While both reviews offer valuable insights into the design space of persuasive social robots, they include very few studies on health behavior change. As Liu et al. noted, until 2019, most research on persuasive social robots focused on abstract decision-making tasks (e.g., desert games), with limited attention to practically meaningful application areas.

More recently, an emerging line of work has focused on social robots as mental health interventions, with several reviews published. For example, Spitale et al. \cite{spitale2022} provided an overview of research on social robots for mental health and well-being, mapping out application contexts and methodologies (e.g., study settings, data collection methods). Similarly, Scoglio et al. \cite{scoglio2019} and Guemghar et al. \cite{guemghar2022} reviewed studies of robots in mental health, summarizing robot types, intervention outcomes, and methodological characteristics. While these reviews partly inform the design and evaluation of persuasive social robots, none focus specifically on the use of social robots to support physical health behavior change. This gap makes the present review timely.

\subsection{Behavioral intervention technology}
Designing behavioral intervention technology is a multifaceted task \cite{moller2017}. According to the Behavioral Intervention Technology Model, the design process involves four constituent parts \cite{mohr2014}. It begins with defining \textit{intervention aims}. For instance, a weight loss intervention may target reducing caloric intake or increasing physical activity. These aims should be as clearly defined as possible to support a focused treatment plan. Once aims are set, designers select appropriate \textit{behavior change strategies} (discussed in 2.3). These strategies are then translated into \textit{technical characteristics}, such as the choice of delivery medium (text, video, audio). Finally, designers specify the \textit{workflow}, determining when and under what conditions the intervention is delivered \cite{mohr2014}. In sum, identifying behavior change strategies serves as a crucial bridge between abstract intervention goals and concrete design elements. Applied to HRI, understanding these strategies provides essential intermediate knowledge for configuring social robots’ behaviors (e.g., verbal and nonverbal expressions) in ways that align with broader health-promoting goals.

\subsection{Behavior change strategies}
In the Behavioral Intervention Technology Model \cite{mohr2014}, a \textit{behavior change strategy} is defined in the same way as a \textit{behavior change technique (BCT)} in health psychology \cite{abraham2008}. A BCT refers to an observable, replicable, and irreducible component of an intervention designed to alter or redirect the processes that regulate behavior \cite{abraham2008}. For example, physical activity interventions can influence behavior through goal setting and self-monitoring \cite{munoz2006,hillsdon2005}, while alcohol reduction programs often make use of behavior substitution \cite{garnett2018behavior}. To establish a shared vocabulary for such components, Michie and colleagues developed a BCT taxonomy that identifies 26 distinct BCTs \cite{abraham2008}. This taxonomy has become a widely used framework for reporting, synthesizing, and designing behavioral interventions in health psychology. However, because it was originally developed for human–human interventions, its direct applicability to technology is limited. Designers of behavioral intervention technologies still face challenges in tailoring these techniques to specific technology applications \cite{kelders2012}.

Alongside the BCT taxonomy, an influential framework specifically developed for designing persuasive technology is the Persuasive Systems Design (PSD) framework, which proposes 28 design principles organized into four categories: primary task support (e.g., self-monitoring), dialogue support (e.g., praise), system credibility support (e.g., authority), and social support (e.g., social comparison) \cite{oinas2018persuasive}. These principles are also regarded as behavior change strategies in the literature \cite{asbjornsen2019,nkwo2021} and are often recognized as useful design guidelines in HCI research \cite{mohadis2016designing}. At the same time, the PSD framework has faced criticism for lacking comprehensiveness \cite{lehto2011,hussian2023} and for including overlapping strategies (e.g., social learning vs. social facilitation) \cite{lehto2013}. 

In this work, we aim to uncover behavior change strategies specifically in HRI to provide direct insights for designing social robot-based behavioral interventions. Given our focus on social robots, which are distinct from conventional persuasive technologies, we do not treat existing frameworks (i.e., the BCT taxonomy or PSD framework) as fixed standards. Instead, we use them as provisional knowledge while inductively identifying distinctive patterns that emerge.

\section{Method}
The review process followed the PRISMA (Preferred Reporting Items for Systematic Reviews and Meta-Analyses) guidelines \cite{moher2009} to ensure methodological rigor and transparency.

\begin{figure*}[hb]
  \centering
  \includegraphics[width=\textwidth]{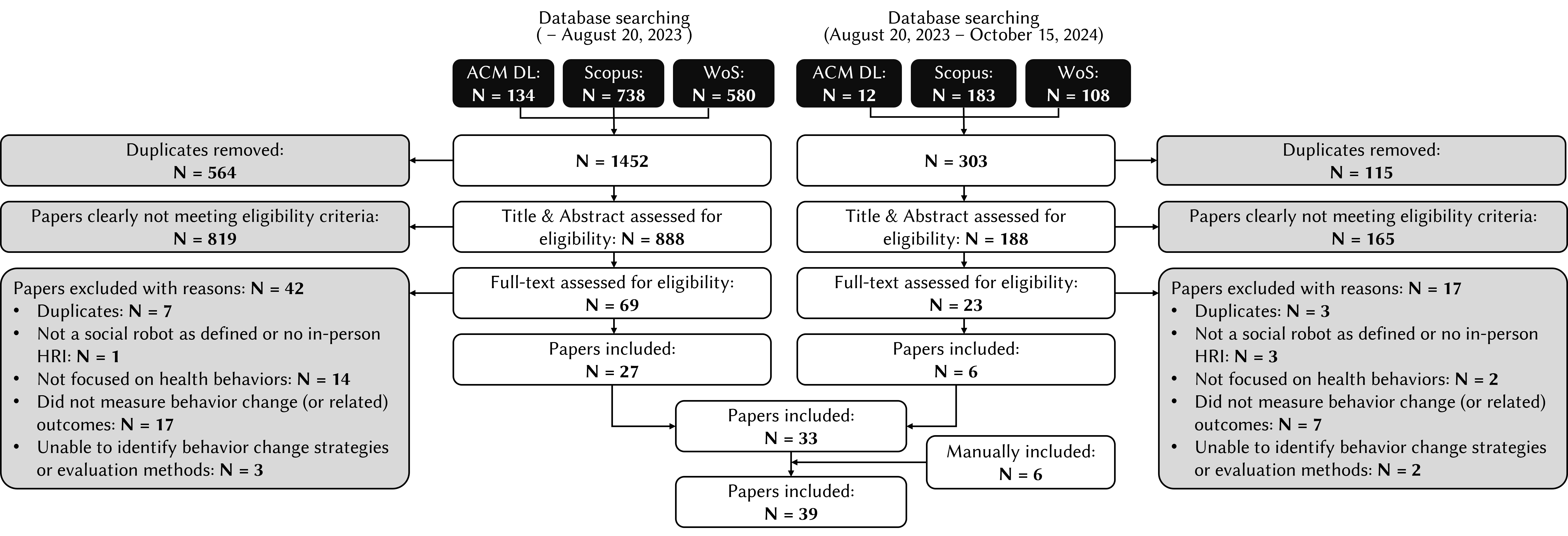}
  \caption{PRISMA flow diagram of included and excluded papers.}
  \Description{The PRISMA flow diagram illustrates the study selection process for this systematic review.}
\end{figure*}

\subsection{Defining terminology}
We defined social robots based on two key criteria. First, following the definition in \cite{naneva2020}, social robots must possess a physical presence. This criterion differentiates social robots from on-screen virtual agents. Second, their core purpose is to engage in \textit{social} interactions with humans \cite{hegel2009}, namely, to act as social actors that interact with people in a verbal and/or nonverbal manner \cite{breazeal2016}. This criterion differentiates social robots from robotic systems whose functions are primarily \textit{non-social}, such as rehabilitation robots that provide physical assistance or industrial robots designed for task-specific operations.

\subsection{Eligibility criteria}
Given our dual focus on both the behavior change strategies and the methods used to evaluate their associated outcomes, we specifically targeted \textit{empirical studies} that both implemented social robots to promote health behavior change and assessed relevant outcomes among human participants.

Guided by this aim, we applied a structured set of inclusion and exclusion criteria using the PICO (Participant, Intervention, Comparator, Outcome) framework, a standard method for structuring eligibility in reviews of behavioral intervention technologies \cite{moher2009}. Eligible studies (\textbf{I}) used social robots (as defined above) as the intervention and involved live human-robot interactions. We excluded studies that used non-social robots (e.g., industrial robots, rehabilitation robots, robotic manipulators, robotic furniture, surgical robots), non-embodied agents (e.g., chatbots), or fully online studies. Exceptions were made for industrial robots equipped with evident social interfaces (e.g., the Baxter robot \cite{1103_fitter2020}). Eligible studies (\textbf{P}) involved data collection from human participants, regardless of demographic characteristics. They were required to (\textbf{O}) measure health behavior change outcomes or outcomes theoretically closely related to behavior change, such as behavior change motivation. Studies focusing exclusively on mental health outcomes, user experience factors, or subjective perceptions of robots were excluded. No criteria were applied to the Comparator (\textbf{C}).

To ensure consistency in publication quality and to maintain a manageable review scope, we included only full research papers published in peer-reviewed journals or conference proceedings and written in English. Publications such as late-breaking reports, student design papers, and work-in-progress papers were excluded. No restrictions were applied regarding publication date.

\subsection{Search strategy}
We developed a keyword set that included both terms related to health behavior change (e.g., “persua*,” “health behavio*,” and “behavio* change”) and synonyms for five specific health behaviors: physical activity (e.g., “exercis*”), dietary behavior (e.g., “eat*”), smoking (e.g., “smok*”), alcohol consumption (e.g., “alcohol*”), and medication adherence (e.g., “medication adherence”). The asterisk automatically truncates terms so that, for example, “behavio*” retrieves behavior, behaviour, behaviors, etc. These keywords were combined with “robot*” and searched within paper titles only. The complete search queries are in the supplementary material.

We selected the Web of Science (WoS) and Scopus as the primary databases due to their extensive coverage across most scientific fields \cite{falagas2008}. The ACM Digital Library (ACM DL) was also included, as it is a leading repository for computer science research \cite{hennessey2012}, and prior evidence indicates that it is not fully indexed in Scopus or WoS \cite{valente2022}. IEEE Xplore was initially considered but excluded because of its substantial overlap with records in Scopus and WoS \cite{valente2022}.

\subsection{Study selection}
The initial search was conducted on 20 August 2023 and yielded 1,452 records across the three databases. After duplicate removal, 888 unique records remained for screening. Screening followed a two-step process. First, titles and abstracts were reviewed against predefined exclusion criteria. At this stage, 819 clearly ineligible records were excluded (e.g., smoke-detection or firefighting robots retrieved using the keywords “smok*” and “cigarette*,” as well as robotic eating aids retrieved using the keyword “eat*”).

Records with unclear eligibility proceeded to full-text screening. During this stage, 42 studies were excluded, leaving 27 eligible papers. All papers excluded at the full-text stage, together with the exclusion reasons, are reported in the supplementary material.

To ensure up-to-date coverage, a follow-up search was conducted on 15 October 2024 using the same search strategy, restricted to publications between 20 August 2023 and 15 October 2024. This search yielded 188 additional unique records, of which six studies met the inclusion criteria after screening. In parallel, a manual hand search was conducted, identifying six relevant papers not captured by the database searches. In total, 39 papers were included in the final analysis.

All screening was conducted independently by the first author, who has expertise in both HRI and behavioral science. The review process was managed using Cadima \cite{kohl2018}, an open-access tool for systematic reviews that supports duplicate detection and screening management. The screening process is summarized in Figure 1.

\subsection{Data extraction and analysis}
We first extracted descriptive information, including publication year, type of social robot, target health behaviors, and methodological characteristics (e.g., study design, setting, duration, and outcome measures for behavior change).

To analyze behavior change strategies, we conducted a thematic analysis using a mixed deductive-inductive coding approach. The BCT taxonomy from health psychology \cite{abraham2008} and the PSD framework \cite{oinas2018persuasive} were used as a priori themes \cite{king2017} to guide our coding. For example, strategies such as \textit{providing feedback on performance} and \textit{providing instruction} were applied directly. At the same time, to capture a broader range of creative ideas, we relied on inductive coding to identify recurring themes that emerged directly from the data. When new patterns appeared outside the predefined themes, such as \textit{leveraging social reciprocity}, they were retained. Finally, all codes were organized into overarching categories. This mixed approach allowed us to balance theory-driven coding with the flexibility to capture insights specific to the HRI context.

The first author conducted data familiarization, coding, code clustering, and theme development using ATLAS.ti software \cite{atlasti}. The second, third, and fourth authors reviewed the coding outputs multiple times and provided feedback for refinement, leading to the finalized results.

\section{Results}
The included 39 studies span 2012–2025. Early work was relatively sparse, with only four studies appearing before 2014 \cite{726_fasola2012, 565_sussenbach2014, 604_baroni2014, 648_broadbent2014}. Five studies were published between 2016 and 2018 \cite{616_swift2016, H1_schneider2016, H3_galvao2018, 518_ritschel2018, 1238_obo2017}. A significant increase occurred from 2019 onward, with 30 studies (77\%) published between 2019 and 2025 \cite{89_winkle2019, 127_rea2021, 163_salomons2022, 194_kharub2022, 221_langedijk2023, 275_okafuji2021, 297_sackl2022, 323_robinson2021, 332_amada2021, 333_fischer2021, 347_fischer2020, 394_da2019, 873_triantafyllidis2023, 886_robinson2020, 894_ren2022, 1041_langedijk2021, 1053_tae2021, 1103_fitter2020, 1105_schneider2021, R2_29_abdulazeem2023, R2_45_jung2023, R2_101_ross2023, H4_lopes2023, 261_ruf2020, R2_81_mayoral2024, R2_108_shao2023, R2_173_zheng2023, H2_giorgi2025, H5_xu2025, H6_casas2021}. This sharp increase reflects the growing recognition of social robots as health behavioral interventions in recent years. A full list of the included papers and the data extraction sheet can be found in the supplementary material.

\subsection{Social robots applied}
Of the 39 included studies, 22 different social robots were used. The most frequently applied was Nao (SoftBank Robotics), used in 12 studies \cite{194_kharub2022,H1_schneider2016, 565_sussenbach2014,604_baroni2014,616_swift2016,886_robinson2020,1105_schneider2021,H3_galvao2018,R2_173_zheng2023,261_ruf2020,H2_giorgi2025,H6_casas2021}. Pepper (SoftBank Robotics) was used in five studies \cite{89_winkle2019,297_sackl2022,R2_101_ross2023,323_robinson2021,R2_108_shao2023}, and Sota (Osaka University and Vstone) in two studies \cite{275_okafuji2021,332_amada2021}. SMOOTH (University of Southern Denmark) was used in \cite{221_langedijk2023,347_fischer2020}. Other robots were each used in a single study: Bandit (University of Southern California) \cite{726_fasola2012}, Beo (Qiron Robotics) \cite{394_da2019}, CommU (Osaka University and Vstone) \cite{1053_tae2021}, Robovie (Advanced Telecommunications Research Institute) \cite{127_rea2021}; JD Humanoid (EZ-Robot) \cite{333_fischer2021}; Socibot Mini (Engineered Arts) \cite{1041_langedijk2021}, Baxter (Rethink Robotics) \cite{1103_fitter2020}, Cozmo (Anki) \cite{873_triantafyllidis2023}, EMYS (EMYS Inc.) \cite{H4_lopes2023}, Keepon (BeatBots) \cite{163_salomons2022}, Sawyer (Rethink Robotics) \cite{R2_29_abdulazeem2023}, Reeti (Robopec) \cite{518_ritschel2018}, PALRO (FUJISOFT Inc.) \cite{1238_obo2017}, IrobiQ (Yujin Robot) \cite{648_broadbent2014}, and Misty (Misty Robotics) \cite{H5_xu2025}. Three studies developed original prototypes \cite{894_ren2022,R2_45_jung2023,R2_81_mayoral2024}. Pictures of all these robots are in the supplementary material.

\subsection{Targeted behavior domains}
Physical activity was most frequently targeted, addressed in 22 of 39 studies (56\%). Most studies focused on light exercises (wrist movements \cite{89_winkle2019}, squats \cite{127_rea2021}, dumbbell lifting \cite{163_salomons2022}, planking \cite{H1_schneider2016}, arm exercise \cite{726_fasola2012, R2_29_abdulazeem2023, 1103_fitter2020, R2_108_shao2023, 1238_obo2017}, or general body movement \cite{R2_81_mayoral2024}). Two studies aimed to reduce sedentary behavior \cite{894_ren2022, H5_xu2025}. Other studies focused on more structured exercises, such as circuit training \cite{297_sackl2022, 616_swift2016}, squash \cite{R2_101_ross2023}, Tai Chi \cite{R2_173_zheng2023}, stationary cycling \cite{394_da2019, 565_sussenbach2014}, treadmill exercise \cite{H6_casas2021}, or mixed exercises \cite{1105_schneider2021, 261_ruf2020}. One study did not specify the activity type \cite{H3_galvao2018}.

Dietary behavior was the second most common focus (9 studies, 23\%), including snacking \cite{886_robinson2020, 275_okafuji2021}, drinking beverage  \cite{194_kharub2022,518_ritschel2018}, eating fruit and vegetable  \cite{604_baroni2014}, and drinking water  \cite{221_langedijk2023, 333_fischer2021, 347_fischer2020, 1041_langedijk2021}. Less common targets included hand‑sanitizing behaviors (2 studies \cite{332_amada2021, 1053_tae2021}), medication (2 studies \cite{648_broadbent2014,H2_giorgi2025}), and drinking alcohol (1 study \cite{R2_45_jung2023}). Three studies targeted multiple health behaviors \cite{323_robinson2021, H4_lopes2023, 873_triantafyllidis2023}. 

\subsection{Behavior change strategies}
Our analysis revealed a rich set of behavior change strategies described across the included studies. These strategies fell into four categories: \textbf{coaching strategies}, \textbf{counseling strategies}, \textbf {social influence strategies}, and \textbf{persuasion-enhancing strategies}, comprising a total of 24 themes (sub-strategies).  

\subsubsection{Coaching strategies}
We identified nine recurring coaching strategies in which robots offered action-level support to assist users in adopting or maintaining target behaviors.

A common strategy was \textbf{providing suggestions}, reported in 15 studies (38\%), where robots offered direct recommendations for healthy actions \cite{221_langedijk2023,275_okafuji2021,332_amada2021,333_fischer2021,347_fischer2020,604_baroni2014,873_triantafyllidis2023,894_ren2022,1041_langedijk2021,518_ritschel2018,648_broadbent2014,R2_45_jung2023,1053_tae2021,H2_giorgi2025,H5_xu2025}. Examples included suggestions like “It would be a good idea to drink some water now \cite{1041_langedijk2021}.” Beyond providing simple suggestions, 19 studies (49\%) reported \textbf{providing instructions}, where robots delivered procedural guidance to help users perform target behaviors. This strategy appeared exclusively in the context of promoting physical activities \cite{127_rea2021,297_sackl2022,394_da2019,565_sussenbach2014,616_swift2016,726_fasola2012,89_winkle2019,894_ren2022,H1_schneider2016,H4_lopes2023,R2_101_ross2023,R2_29_abdulazeem2023,1105_schneider2021,1238_obo2017,163_salomons2022,261_ruf2020,R2_108_shao2023,R2_173_zheng2023,1103_fitter2020}. 

Another frequently reported strategy was \textbf{providing feedback on performance} (23 studies, 59\%), which took three main forms. First, \textit{descriptive feedback}, reported in eight studies \cite{394_da2019,565_sussenbach2014,886_robinson2020,H4_lopes2023,R2_101_ross2023,1238_obo2017,297_sackl2022,726_fasola2012}, involved robots objectively reporting user performance without interpretation or evaluation (e.g., “You’ve cycled for 2 minutes” \cite{394_da2019}). Six mentioned \textit {corrective feedback} \cite{565_sussenbach2014,726_fasola2012,R2_101_ross2023,163_salomons2022,R2_45_jung2023,333_fischer2021}, where robots pointed out mistakes or deviations and guided users toward improvements (e.g., “Raise your left arm higher” \cite{726_fasola2012}). The most common form, \textit {positive motivational feedback} (i.e., praise), appeared in 20 studies \cite{89_winkle2019,127_rea2021,H1_schneider2016,726_fasola2012,297_sackl2022,565_sussenbach2014,604_baroni2014,616_swift2016,394_da2019,R2_29_abdulazeem2023,R2_101_ross2023,873_triantafyllidis2023,R2_45_jung2023,H3_galvao2018,323_robinson2021,163_salomons2022,R2_108_shao2023,R2_173_zheng2023,1238_obo2017,H6_casas2021}, where robots responded positively to desirable behavior (e.g., “You are doing great” \cite{H1_schneider2016}). In addition, two studies reported \textit {negative motivational feedback}: \cite{127_rea2021} used “face-threatening” feedback such as “Can you even make it?”, while \cite{R2_101_ross2023} applied a “scold” strategy (e.g., “That was a bad one”).

Twelve studies (31\%) mentioned the strategy of \textbf{explaining health consequences}, helping users understand why behavior change was important. In ten of them \cite{221_langedijk2023,333_fischer2021,616_swift2016,873_triantafyllidis2023,H4_lopes2023,1041_langedijk2021,1053_tae2021,1238_obo2017,518_ritschel2018,H5_xu2025}, robots emphasized the \textit{benefits of healthy behaviors} (e.g., “Drinking water is crucial for blood and bones \cite {1041_langedijk2021}”). In addition, four of them \cite{323_robinson2021, H4_lopes2023,194_kharub2022,1238_obo2017} described robots conveying \textit {risks of unhealthy behaviors} (e.g., “High caffeine intake can lead to anxiety and higher blood pressure \cite{194_kharub2022}”).  

Another frequently reported strategy was \textbf{behavioral modeling} (13 studies, 33\%), where robots physically demonstrated target behaviors \cite{297_sackl2022,616_swift2016,726_fasola2012,H1_schneider2016,H4_lopes2023,R2_101_ross2023,1105_schneider2021,1103_fitter2020,1238_obo2017,261_ruf2020,R2_108_shao2023,R2_173_zheng2023,R2_81_mayoral2024}, such as showing arm exercises for users to follow \cite{726_fasola2012}. 

Five studies (13\%) mentioned \textbf {prompting goal setting}, in which robots asked users to set a concrete goal to achieve \cite{873_triantafyllidis2023,886_robinson2020,H4_lopes2023,604_baroni2014,H3_galvao2018}, such as eating N portions of fruits next week \cite{604_baroni2014}. 

Three studies (8\%) mentioned the strategy of \textbf {gamifying health activities}. This included \textit{embedding creative activities}, for example, \cite{1103_fitter2020} introduced an exercise game that transformed users' physical movements into musical compositions, and \textit{leveling up the challenge}, where robots gradually guided users toward more challenging and complex activities \cite{726_fasola2012,1103_fitter2020,R2_108_shao2023}.

Two studies (5\%) described the strategy of \textbf{providing interactive rewards}, where the robot rewarded users for performing a desired behavior. In one study, the robot sang, danced, or invited users to give it a fist bump \cite{873_triantafyllidis2023}; in another, the robot used light effects, sounds, and bubbles (produced by a geared motor with bubble solution attached to the robot) \cite{R2_81_mayoral2024}.

Two studies (5\%) mentioned using robots for \textbf{enabling self-monitoring of behavior}, such as providing a digital calendar for users to log their behaviors \cite{H4_lopes2023} or asking them to track progress in diaries \cite{886_robinson2020}.

\subsubsection{Counseling strategies}
Another category is counseling strategies, where robots helped users find their internal motivation for behavior change. Unlike coaching strategies that focus on actions, counseling strategies prompt users to think and reflect. Within this category, four strategies were identified across three studies. 

Two studies (5\%) mentioned \textbf{prompting verbalization of pros and cons}, where the robot encouraged users to articulate the advantages and disadvantages of their current behaviors to surface internal conflicts and values, such as by asking “Would anything else concern you if you keep doing things the way you are doing them?” \cite{H4_lopes2023, H3_galvao2018}. Three studies (8\%) mentioned \textbf{prompting reflection on past success}, where robots guided users to recall previous achievements or successful behavior changes, such as by asking “Tell me about a time in your life when you had been doing better than you are now” \cite{H4_lopes2023,886_robinson2020, H3_galvao2018}. Two studies (5\%) mentioned \textbf {prompting imagination of future outcomes}, where the robot guided users to envision future scenarios, such as achieving goals or overcoming challenges \cite{H3_galvao2018,886_robinson2020}. Additionally, two studies (5\%) mentioned \textbf {problem solving}, where the robot helped users to generate coping plans to address barriers to behavior change \cite{H3_galvao2018, H4_lopes2023}. 

\subsubsection{Social influence strategies}
Social influence strategies were frequently reported, in which robots were designed to influence users by leveraging social roles, norms, or group dynamics.

Six studies (15\%) employed \textbf{shaping robot authority}, either by \textit{emphasizing expert identity} (e.g., “I have been programmed by physiotherapists who specialize in exercise” \cite{89_winkle2019}) or by \textit{referring to scientific evidence} (e.g., “Studies indicate that dehydration can cause fatigue” \cite{221_langedijk2023,275_okafuji2021,194_kharub2022,616_swift2016,H5_xu2025}).

Three studies (8\%) mentioned \textbf{invoking social conformity}, where robots tried to motivate users by suggesting that a behavior was common among others. For example, robots said, “Most people have taken something to drink” \cite{221_langedijk2023,347_fischer2020} or “My recommendation is based on the preferences of everyone on this floor” \cite{275_okafuji2021}.

Three studies (8\%) touched on the idea of \textbf {leveraging social reciprocity}, where robots framed behavior change as part of a mutual exchange. In \cite{894_ren2022}, a robot first asked users to complete a voluntary task (e.g., returning books) and then provided exercise instructions as a reciprocal favor. In \cite{726_fasola2012,1103_fitter2020}, robots instructed participants to exercise while also requesting to be taught in return.

Two studies (5\%) focused on \textbf{cueing social rituals}, using those symbolic or culturally meaningful patterns of interaction as a way of persuasion. For instance, robots initiated a “cheers” gesture to promote drinking water \cite{R2_45_jung2023,347_fischer2020}.

Two studies (5\%) mentioned \textbf{mediating social learning}. In \cite{H4_lopes2023}, the robot provided information about other people's health behaviors to allow users to observe and learn, and in \cite{R2_108_shao2023}, a robot moderated group exercise sessions to promote mutual learning. 

Two studies (5\%) mentioned \textbf{mediating social support}. In \cite{H4_lopes2023}, the robot prompted users in the same intervention to encourage each other to complete health activities, and in \cite{648_broadbent2014}, the robot contacted caregivers via Skype when users missed medication doses. 

Other strategies were reported less frequently. One study (3\%) described \textbf{enabling social comparison}. In this study, users exercised alongside a robot whose performance was significantly better than that of the users, thereby motivating greater effort  \cite{H1_schneider2016}. Another study described \textbf{enabling cooperation}, in which the robot and users worked together to achieve a shared objective. In this case, they played a handclap game requiring coordinated movements to complete the task \cite{1103_fitter2020}.

\subsubsection{Persuasion-enhancing strategies} 
We also identified a set of strategies that do not directly persuade on their own but are often used alongside other strategies. We term these persuasion‑enhancing strategies.

One common strategy, reported in 13 studies (34\%), was \textbf{enhancing embodied presence}, with robots using multi-modal embodied cues to appear more socially present. This included \textit{increasing proximity}, where the robot tried to be physically closer to users while initiating persuasion \cite{604_baroni2014,332_amada2021,894_ren2022, R2_45_jung2023}. Seven studies mentioned \textit{leveraging gestures} \cite{323_robinson2021,333_fischer2021,604_baroni2014,163_salomons2022,R2_45_jung2023,297_sackl2022,616_swift2016}. For example, in \cite{333_fischer2021}, a robot gestured toward a glass of water while suggesting users to drink, while in \cite{163_salomons2022}, a robot rhythmically moved its body while instructing the participant to perform dumbbell exercises. Five studies mentioned \textit{establishing eye contact}, where robots maintained gaze with users during persuasion \cite{604_baroni2014,332_amada2021,347_fischer2020,323_robinson2021,518_ritschel2018}. Two study described \textit{animating facial expressions} \cite{R2_29_abdulazeem2023,518_ritschel2018}, and another experimented with \textit{using multiple robots} instead of a single robot to persuade \cite{1053_tae2021}. 

Another strategy, mentioned in nine studies (24\%), was \textbf{personalizing persuasion}, where robots adapted their persuasive content to the unique characteristics of individual users. Personalization was implemented in several ways. Some studies described tailoring to demographic characteristics \cite{221_langedijk2023,616_swift2016,873_triantafyllidis2023,R2_101_ross2023,1103_fitter2020}, such as adjusting instructions to match the physical and cognitive abilities of older adults \cite{1103_fitter2020}. Others personalized based on users’ behavioral history, with robots delivering feedback or suggestions informed by past performance \cite{726_fasola2012,873_triantafyllidis2023,H4_lopes2023,323_robinson2021}. Personalization was also achieved by adjusting to users’ motivational status \cite{R2_101_ross2023} or their dynamic preferences \cite{1105_schneider2021,873_triantafyllidis2023}.

Another strategy, mentioned in nine studies (24\%), was \textbf{stimulating social bonding}, involving robots using relational behaviors to create a sense of connection with users. Five studies described robots \textit{calling users by name} to initiate relationship building \cite{726_fasola2012,873_triantafyllidis2023,H4_lopes2023,604_baroni2014,648_broadbent2014}. Five studies reported robots \textit{showing empathy} by actively inquiring about users’ feelings \cite{1041_langedijk2021,89_winkle2019}, being happy when users made progress \cite{604_baroni2014}, showing concern when users were tired \cite{R2_108_shao2023}, or offering reassurance when users failed \cite{726_fasola2012}. Three studies mentioned \textit{signaling common ground}, where the robot talked about shared contexts \cite{333_fischer2021} or previous interactions \cite{726_fasola2012,1041_langedijk2021}. Three studies mentioned \textit{using humor} to induce positive affectivity \cite{347_fischer2020, 194_kharub2022,221_langedijk2023}, and one study reported \textit{demonstrating similarity}, where the robot claimed that it had shared preferences with its users (e.g., “It seems like we have similar ideas about exercising”  \cite{89_winkle2019}).

\subsection{Evaluation methods}
\subsubsection{Study designs}
Nineteen studies (49\%) aimed to compare specific design factors of robots \cite{H5_xu2025,H1_schneider2016,333_fischer2021,332_amada2021,275_okafuji2021,194_kharub2022,1053_tae2021,R2_29_abdulazeem2023,604_baroni2014,89_winkle2019,H2_giorgi2025,R2_108_shao2023,1105_schneider2021,R2_101_ross2023,1041_langedijk2021,1103_fitter2020,616_swift2016,726_fasola2012,127_rea2021}. Twelve (31\%) aimed to examine the overall effects of robots by comparing them with alternative interventions: Some compared robots with humans \cite{894_ren2022}, virtual agents \cite{163_salomons2022}, text displays \cite{565_sussenbach2014}, videos \cite{261_ruf2020}, or control conditions without robots \cite{R2_81_mayoral2024,394_da2019, H6_casas2021}, and some adopted a single-group repeated-measures design \cite{873_triantafyllidis2023,H4_lopes2023,323_robinson2021} or randomized controlled trials with repeated measures to evaluate overall robot effects \cite{648_broadbent2014,886_robinson2020}. Eight (20\%) were exploratory studies that descriptively observed robot effects without comparison groups \cite{297_sackl2022,347_fischer2020,H3_galvao2018,1238_obo2017,518_ritschel2018,R2_173_zheng2023,R2_45_jung2023,221_langedijk2023}.

\subsubsection{Study settings}
Of the 39 studies, 18 (46\%) conducted evaluations in labs \cite{127_rea2021,194_kharub2022,333_fischer2021,394_da2019,565_sussenbach2014,616_swift2016,886_robinson2020,89_winkle2019,H1_schneider2016,H3_galvao2018,R2_29_abdulazeem2023,1041_langedijk2021,1105_schneider2021,R2_45_jung2023,1103_fitter2020,R2_173_zheng2023,H2_giorgi2025,H5_xu2025}. Twenty (51\%) were carried out in field settings, including public spaces \cite{221_langedijk2023,347_fischer2020,332_amada2021,1053_tae2021,518_ritschel2018}, user homes \cite{163_salomons2022,261_ruf2020}, workplaces \cite{894_ren2022,275_okafuji2021}, senior living facilities \cite{1238_obo2017,648_broadbent2014,R2_108_shao2023,726_fasola2012}, schools \cite{873_triantafyllidis2023,604_baroni2014}, gyms \cite{297_sackl2022,R2_101_ross2023}, daycare centers \cite{R2_81_mayoral2024}, health clinics \cite{323_robinson2021}, and rehabilitation centers \cite{H6_casas2021}. One study was conducted online \cite{H4_lopes2023} (exceptionally included due to high relevance).

\subsubsection{Participant characteristics}
Nineteen (49\%) studies conducted evaluation in general adults \cite{275_okafuji2021,297_sackl2022,333_fischer2021,565_sussenbach2014,886_robinson2020,89_winkle2019,H3_galvao2018,H4_lopes2023,R2_101_ross2023,163_salomons2022,R2_45_jung2023,127_rea2021} or students \cite{394_da2019,894_ren2022,H1_schneider2016,1041_langedijk2021,1105_schneider2021,194_kharub2022,H5_xu2025}. Nine studies (23\%) targeted older adults \cite{726_fasola2012,R2_29_abdulazeem2023,1238_obo2017,261_ruf2020,648_broadbent2014,R2_108_shao2023,R2_173_zheng2023,1103_fitter2020,H2_giorgi2025}, and four studies (10\%) targeted children \cite{873_triantafyllidis2023, 604_baroni2014, R2_81_mayoral2024} or adolescents \cite{616_swift2016}. Two studies (5\%) focused on clinical patients \cite{323_robinson2021,H6_casas2021} and five (13\%) engaged with some random public \cite{221_langedijk2023,332_amada2021,347_fischer2020,1053_tae2021,518_ritschel2018}. 

\subsubsection{Intervention duration}
78\% of lab studies (14 out of 18) were one-time interaction \cite{R2_45_jung2023, R2_173_zheng2023, 1105_schneider2021, 1041_langedijk2021, R2_29_abdulazeem2023, H3_galvao2018, H1_schneider2016, 89_winkle2019, 394_da2019, 333_fischer2021, 1103_fitter2020, 194_kharub2022, 127_rea2021, H2_giorgi2025}. The remaining four employed extended interventions, including 2 weekly sessions \cite{H5_xu2025}, three sessions over eight weeks \cite{886_robinson2020}, four sessions over two weeks \cite{616_swift2016}, and 18 daily sessions \cite{565_sussenbach2014}. Among field studies, 55\% (11 out of 20) were one-time interaction \cite{894_ren2022,873_triantafyllidis2023,726_fasola2012,604_baroni2014,347_fischer2020,323_robinson2021,297_sackl2022,518_ritschel2018,221_langedijk2023,1053_tae2021,332_amada2021}. The remaining nine implemented multiple sessions, including three sessions in two days \cite{R2_101_ross2023}, six daily sessions \cite{1238_obo2017}, seven daily sessions \cite{261_ruf2020}, 14 daily sessions \cite{163_salomons2022}, a three‑week intervention \cite{275_okafuji2021}, a six‑week intervention \cite{648_broadbent2014}, eight weekly sessions \cite{R2_81_mayoral2024}, 16 sessions in two months \cite{R2_108_shao2023}, and 119 sessions between three and six months \cite{H6_casas2021}.

\subsubsection{Behavior change outcome measures}
First, \textbf{observational behavioral measures} were used in four studies (10\%). In these cases, researchers directly observed participants during interaction sessions and manually noted what they did, for example, counting how many exercise repetitions participants completed \cite{89_winkle2019, 127_rea2021} or how much water they drank from cups \cite{333_fischer2021, 1041_langedijk2021}. 

Second, \textbf{video-coded behavioral measures} were used in 12 studies (31\%) \cite{1103_fitter2020,221_langedijk2023,275_okafuji2021,332_amada2021,347_fischer2020,565_sussenbach2014,R2_29_abdulazeem2023,R2_45_jung2023,1053_tae2021,R2_173_zheng2023,R2_81_mayoral2024,H5_xu2025}. Here, researchers recorded sessions and later coded participants’ behavior, such as how much time they spent exercising \cite{H5_xu2025}. 

Third, \textbf{sensor-based behavioral measures} were reported in 13 studies (33\%) \cite{394_da2019,565_sussenbach2014,616_swift2016,726_fasola2012,H1_schneider2016,R2_101_ross2023,1238_obo2017,163_salomons2022,518_ritschel2018,R2_108_shao2023,R2_81_mayoral2024,1103_fitter2020,H6_casas2021}. These measures relied on sensors embedded in the robot or on external sensing devices (e.g., accelerometers) to collect behavioral data.

Fourth, eight (21\%) used \textbf{self-report behavioral measures}. Among these studies, six used behavioral assessment scales, such as the Medication Adherence Report Scale \cite{648_broadbent2014}, the Active Australia Survey \cite{323_robinson2021}, Borg Scale of Effort \cite{H1_schneider2016, 1103_fitter2020, 127_rea2021}, or other study-specific scales \cite{873_triantafyllidis2023}. Another asked participants to keep diaries of snack and drink consumption \cite{886_robinson2020}, and one simply asked whether participants had exercised more in the past week \cite{H3_galvao2018}. 

Next, \textbf{self-report motivational measures} were used in 20 studies (47\%). Five measured intrinsic motivation with instruments such as the Intrinsic Motivation Inventory \cite{894_ren2022, R2_101_ross2023, 297_sackl2022} and the Exercise Self-Regulation Questionnaire \cite{616_swift2016}, and some used unvalidated scales \cite{726_fasola2012}. Three measured self-efficacy, using tools like the Sports Efficacy for Exercise Questionnaire \cite{127_rea2021} or other unspecified scales \cite{H4_lopes2023,1103_fitter2020}. Three used other motivation measurement like the Motivational Thought Frequency Scale \cite{886_robinson2020}, NASA-TLX \cite{565_sussenbach2014}, or other \cite{394_da2019}. Seven used single Likert-type questions \cite{333_fischer2021, 1105_schneider2021, 163_salomons2022, 323_robinson2021, H4_lopes2023, 194_kharub2022, H2_giorgi2025}, such as “How much more motivated are you to change a health behavior? \cite{323_robinson2021}” One study asked participants to make a promise before and after interacting with a robot (e.g., “I promise to eat N portions of fruit”) and then checked whether their commitment increased \cite{604_baroni2014}. Two studies measured motivation qualitatively, asking open questions such as “Would you like to keep up the daily exercise with the robot after this experiment?” \cite{1238_obo2017} or “Was it motivating for you to train with the robot?” \cite{261_ruf2020}. 

Finally, \textbf{physiological measures} were reported in four studies (10\%). Three measured indicators such as heart rate \cite{616_swift2016, 565_sussenbach2014, H6_casas2021}, and one tracked health outcomes, such as body mass index \cite{886_robinson2020}.


\section{Discussion}
This review examined the behavior change strategies applied in health-promoting social robots and the ways in which behavior change outcomes have been evaluated. Below, we summarize the key findings and outline potential directions for future research.

\subsection{Principal findings}
\subsubsection{Behavior change strategies (RQ1)} 
Our analysis shows that social robots are most commonly designed to deliver coaching strategies, such as providing instructions and feedback, modeling behaviors, and rewarding users. While these strategies are also prevalent in other persuasive technologies, their implementation in HRI often takes richer forms. For example, in behavioral modeling, robots can physically demonstrate actions for users to follow, achieving a level of realism that on-screen agents cannot easily replicate. Similarly, reward strategies in HRI extend beyond the textual praise or virtual rewards typically used in conventional digital interventions \cite{lewis2016}, and may include tangible rewards \cite{R2_81_mayoral2024} or physical interactions \cite{873_triantafyllidis2023}. Together, these approaches expand traditional notions of digital coaching and highlight the potential of HRI to create more immersive coaching experiences.

At the same time, we observed that some well-established coaching strategies are rarely mentioned in HRI. Two notable examples are goal setting and self-monitoring, which are widely used in app- and web-based systems \cite{roberts2017,stockwell2019,wang2019health,aldenaini2020,ndulue2022games}. A likely explanation is technical: social robots often lack built-in displays that allow users to set goals, log progress, or directly view outcomes. Their physical presence also makes them less suitable for 24/7 tracking than portable devices. Looking ahead, HRI designers should explore ways to enable goal setting and self-monitoring if robots are to support long-term health behavior change. One potential direction is to integrate social robots into broader digital health ecosystems \cite{stanojevic12digital}. In such systems, other digital health platforms (e.g., smartphones or wearables) may handle progress tracking, while the robot focuses on social interaction, as demonstrated in \cite{H6_casas2021}.

Additionally, we found that nearly all existing behavior change strategies in HRI are positively framed. This pattern appears to reflect an implicit assumption that robots should always be polite, encouraging, and quick to offer praise or rewards. However, one study included in this review suggests that negatively valenced strategies can also effectively motivate behavior change \cite{127_rea2021}. In the broader behavior change literature, strategies such as negative reinforcement, fear appeals, or punishment have been shown to be powerful in certain intervention contexts (e.g., smoking cessation \cite{sutton1984}). Examining how such negatively valenced strategies, when delivered by robots, influence users may therefore represent a possible direction for future research.

Furthermore, we identified a particularly rich set of social influence strategies, many of which extend beyond those captured in existing BCTs or the PSD framework. These include invoking social conformity, reciprocity, and social rituals. Some common strategies, such as social comparison and cooperation, also appeared, but in newer forms. Unlike conventional persuasive technologies, which primarily mediate social influence between people, robots can act as direct sources of influence. They participate in health activities, cooperate with users \cite{1103_fitter2020}, and invite direct performance comparisons \cite{H1_schneider2016}. Together, these patterns point to an expanded social influence design space enabled by the embodied nature of social robots. To further develop this space, future HRI research could draw more extensively on insights from social psychology. For example, commitment-based strategies (i.e., encouraging users to make and maintain commitments to others), as described in Cialdini’s principles of social influence \cite{cialdini2009}, appear to remain unexplored in HRI.

Another noteworthy trend we identified is the strengthening of social bonds between users and robots as a means of enhancing persuasion. A similar idea has been discussed in chatbot research, which suggests that effective interventions depend not only on a system’s persuasive capabilities but also on its ability to build relationships with users \cite{zhang2020}. For social robotic interventions, this relational capacity may be even more critical. Because social robots are more readily perceived as social entities, users may expect more affective and socially meaningful interactions with them \cite{xu2025does}. Accordingly, an important next step for research on persuasive social robots is to examine how persuasive strategies can be expanded by adopting a relational perspective. Insights from relationship science are particularly informative in this regard. Prior work shows that intimate compliance strategies—such as expressions of caring (e.g., “Your health really matters to me”) and liking (e.g., “You would look great if you were in good shape”)—conveyed by close partners can profoundly influence health behaviors \cite{dennis2006, dennis2011social}. Investigating how such strategies could play a role in HRI may therefore represent a promising research direction.

\subsubsection{Evaluation methods (RQ2)} 
Methodologically, we found that more than half of the reviewed studies have moved from laboratory to field settings, aiming to evaluate behavior change in more ecologically valid contexts. This shift represents an encouraging trend in the field. However, we also observed that social robot interventions are rarely evaluated longitudinally. Most studies still rely on single-interaction or short-term designs, making it difficult to determine if observed effects reflect genuine persuasive impact or are primarily driven by novelty effects \cite{221_langedijk2023,333_fischer2021, R2_81_mayoral2024}. This pronounced lack of long-term evaluation stands in contrast to prior reviews of other behavioral intervention technologies (e.g., chatbots \cite{aggarwal2023,wang2025}), which report a substantially higher proportion of longitudinal assessments.

This discrepancy is understandable, as implementing social robots in long-term interventions and data collection poses greater technical and ethical challenges than conventional digital platforms. From a technical perspective, social robots are prone to hardware failures, sensing instability, and maintenance issues, particularly when deployed in unsupervised field environments  \cite{261_ruf2020}. From an ethical perspective, long-term behavioral monitoring via embodied robots may raise concerns related to physical intrusiveness \cite{melenhorst2004}, deprivation of autonomy \cite{peng2025}, and violations of privacy \cite{peng2025,von2011living}, which further constrain sustained data collection. These challenges help explain why the majority of existing studies in this domain remain limited to short-term feasibility testing. 

To enable more robust long-term empirical research in this domain, future studies need to address these data collection challenges more explicitly. One possible approach is to place greater emphasis on multiple complementary measures, such as users’ self-reported behavioral data, thereby reducing dependence on the robot’s sensing reliability and continuous monitoring. A second possible direction is to move beyond purely statistical analytic approaches and incorporate more micro-level qualitative methods. For example, in one of the included studies \cite{565_sussenbach2014}, the researchers analyzed detailed interactional patterns in video data from a smaller sample over time, yielding in-depth insights into how robots influence users’ behavior. Such approaches may help mitigate some of the technical and ethical risks associated with large-scale participant recruitment while still producing meaningful empirical evidence.

Additionally, we found that the way motivation is currently measured is problematic. Many studies rely on single-item, loosely defined measures, while theoretically grounded motivational constructs remain underutilized. For example, the Theory of Planned Behavior identifies \textit{behavioral intention} (i.e., desire to perform a behavior) as the proximal determinant of action \cite{ajzen1991}. The Transtheoretical Model conceptualizes motivation as different stages of \textit{readiness to change} \cite{prochaska1983stages}. Self-Determination Theory defines motivation along a continuum of autonomy, ranging from controlled forms of \textit{extrinsic motivation} to more self-determined \textit{intrinsic motivation} \cite{deci2000}. While these constructs all relate to motivation, they respond differently to various behavior change strategies and show distinct associations with behavioral outcomes \cite{knittle2018}. Future studies should therefore avoid treating motivation as a one-size-fits-all concept and instead ground their measures in clear theoretical frameworks to improve the interpretability and comparability of results.

\subsection{Ethical considerations}
Although this review identifies a range of potential strategies through which social robots can influence human behavior, their future application in HRI research must be accompanied by careful ethical consideration. Persuasion via social robots inherently operates along a delicate boundary between being supportive and being coercive \cite{halttu2022}. As Fogg argues, increasing the “social volume” of a persuasive technology raises the stakes of its influence, meaning it can either “win bigger or lose bigger” \cite{fogg2002}. Owing to their human-like embodiment and socially evocative nature, social robots may indeed enhance persuasive effectiveness, but they can also amplify the potential harms of inappropriately applied strategies.

These concerns may be particularly salient for strategies such as emphasizing health risks, delivering negative feedback, or invoking social comparison. While such strategies have been shown to elicit compliance in certain contexts \cite{127_rea2021,194_kharub2022}, their repeated or prolonged use may also evoke feelings of guilt, patronization, or demoralization \cite{orji2019socially}. Over time, such negative experiences may undermine users’ well-being. We therefore urge HRI researchers and designers to apply persuasive strategies with caution and, where possible, to complement compliance-focused evaluations with assessments of potential adverse effects, such as psychological reactance \cite{dillard2005}. Such efforts are essential for delineating the ethical boundaries of persuasive design in social robots.

\subsection{Limitations of this review}
The first limitation of this review is that we did not assess the effectiveness of the identified strategies. As our primary aim was to map design and evaluation practices as comprehensively as possible, we deliberately included studies spanning a wide range of behavioral domains (e.g., exercise, diet, alcohol use), outcome measures (e.g., objective and subjective), and study designs (e.g., comparisons between strategies or between systems). Given this substantial methodological heterogeneity, drawing conclusions about the relative effectiveness of specific strategies was not feasible. We hope that this review can instead serve as a foundation for more standardized future research, which would enable such analyses.

A second limitation concerns the screening process. Paper screening was conducted by the first author only. Although the eligibility criteria were jointly developed by the full research team, relying on a single screener introduces the possibility of selection bias. 

Finally, despite our efforts to conduct a comprehensive search, some relevant work may have been missed. For example, we identified several design-focused papers that did not include empirical evaluations (e.g., \cite{antony2023,looije2017}) and ultimately excluded them to keep the scope of the review manageable. Readers are encouraged to consult these additional sources for complementary perspectives on the design of persuasive social robots.

\section{Conclusion}
To our knowledge, this work represents the first systematic attempt to map behavior change strategies and evaluation methods in the context of social robots for health behavior change. Across 39 empirical studies, we found that HRI researchers have explored a wide range of coaching, counseling, and social influence strategies, and have leveraged robots’ embodied presence, personalization, and capacity for social bonding to enhance persuasive impact. On the evaluation side, we reviewed commonly used approaches for assessing behavior change outcomes and highlighted recurring limitations, including the scarcity of longitudinal studies and the limited theoretical grounding of many motivational measures. For the HRI community, we hope that this review provides an initial shared knowledge base to support more theoretically grounded, methodologically robust, and ethically informed design and evaluation of persuasive social robots in health-promoting contexts.

\begin{acks}
The authors thank the anonymous reviewers for their constructive feedback. The first author also thanks Zhuochao Peng for valuable guidance on the thematic analysis. This work was supported by the China Scholarship Council (Grant No. 202207720098). ChatGPT 5.2 was used for proofreading. All content was created by the authors themselves. 
\end{acks}

\bibliographystyle{ACM-Reference-Format}
\bibliography{HRI2026REFERENCE}

\end{document}